\newcommand{\vc}[1]{\mathbf{#1}} 					
\newcommand{\vect}[1]{\mathbf{#1}} 					
\newcommand{\cv}[0]{\mathbf{c}} 					
 
\newcommand{\lone}{$\ell1$}

\newcommand{\Expect}{{\rm I\kern-.3em E}}				

\newcommand{\com}{\vect{c}}
\newcommand{\dcom}{\vect{\dot{c}}}
\newcommand{\ddcom}{\vect{\ddot{c}}}
\newcommand{\state}{\vect{S}}

\documentclass[letterpaper, 10 pt, journal, twoside]{IEEEtran}

\usepackage{graphics} 
\usepackage{color, colortbl}
\usepackage[ruled,linesnumbered]{algorithm2e}
\usepackage{epsfig} 
\usepackage{amsmath} 
\usepackage{amssymb}  
\usepackage{bm}
\usepackage{booktabs}
\usepackage{tabularx}
\usepackage{multirow}
\usepackage{graphicx,enumitem}
\usepackage{xcolor}
\usepackage{soul}
\usepackage{leftidx}

\ifCLASSINFOpdf
\else
\fi
\begin{document}
%
\title{Solving Footstep Planning as a Feasibility Problem using L1-norm Minimization (Extended Version)}

\author{Daeun Song$^{1}$, 
        Pierre Fernbach$^{2}$, 
        Thomas Flayols$^{2}$, 
        Andrea Del Prete$^{3}$, 
        Nicolas Mansard$^{2}$ \\
        Steve Tonneau$^{4}$, 
        and Young J. Kim$^{1}$ 
\thanks{\textit{Corresponding author: Young J. Kim.}}
\thanks{$^{1}$D. Song and Y. J. Kim are with the Department of Computer Science and Engineering, Ewha Womans University, Seoul 03760, South Korea
    {\tt\small daeunsong@ewhain.net, kimy@ewha.ac.kr}}
\thanks{$^{2}$P. Fernbach, T. Flayols and N. Mansard are with the CNRS, LAAS, Universit\'e de Toulouse, Toulouse 31400, France
    {\tt\small pfernbach@laas.fr, tflayols@laas.fr, nmansard@laas.fr}}
\thanks{$^{3}$A. Del Prete is with the Department of Industrial Engineering, University of Trento, Via Sommarive 9, Trento 30123, Italy
    {\tt\small andrea.delprete@unitn.it}}
\thanks{$^{4}$S. Tonneau is with the University of Edinburgh, IPAB, Edinburgh EH8 9YL, U.K.
    {\tt\small stonneau@ed.ac.uk}}

}

\newcommand{\rom}[1]{\romannumeral #1)}

%
%

\markboth{Extended version of IEEE Robotics and Automation Letters accepted paper, 2021}
{Song \MakeLowercase{\textit{et al.}}: Solving Footstep Planning as a Feasibility Problem using L1-norm minimization} 

%



\maketitle

\begin{abstract}
One challenge of legged locomotion on uneven terrains is to deal with both the {\em discrete problem} of selecting a contact surface for each footstep and the {\em continuous problem} of placing each footstep on the selected surface. Consequently, footstep planning can be addressed with a Mixed Integer Program (MIP), an elegant but computationally demanding method, which can make it unsuitable for online planning. 
We reformulate the MIP into a cardinality problem, then approximate it as a computationally efficient \lone-norm minimisation, called SL1M. Moreover, we improve the performance and convergence of SL1M by combining it with a sampling-based root trajectory planner to prune irrelevant surface candidates.

Our tests on the humanoid Talos in four representative scenarios show that SL1M always converges faster than MIP. For scenarios when the combinatorial complexity is small ($<10$ surfaces per step), SL1M converges at least two times faster than MIP with no need for pruning.
In more complex cases, SL1M converges up to 100 times faster than MIP with the help of pruning. Moreover, pruning can also improve the MIP computation time. The versatility of the framework is shown with additional tests on the quadruped robot ANYmal.
\end{abstract}

\begin{IEEEkeywords}
Humanoid and Bipedal Locomotion, Legged Robots, Motion and Path Planning
\end{IEEEkeywords}

%
\IEEEpeerreviewmaketitle

\section{Introduction}
%
%
%
%

\IEEEPARstart{F}{ootstep} planning consists of computing a sequence of footstep positions on which a legged robot should step to reach a desired goal position. It is thus a crucial problem for legged locomotion.

Footstep planning can be characterised by its combinatorial aspect. The parts of the environment where the footsteps can be placed (contact surfaces) are often disjoint. Thus, a discrete choice of a surface must be made. This choice has an impact on all future footstep locations, as they are constrained relative to each other by non-linear kino-dynamic constraints. This means that the discrete choices of contact surfaces must be considered simultaneously for all footsteps.

\begin{figure}[tb]
\centering
\includegraphics[width=9.0cm]{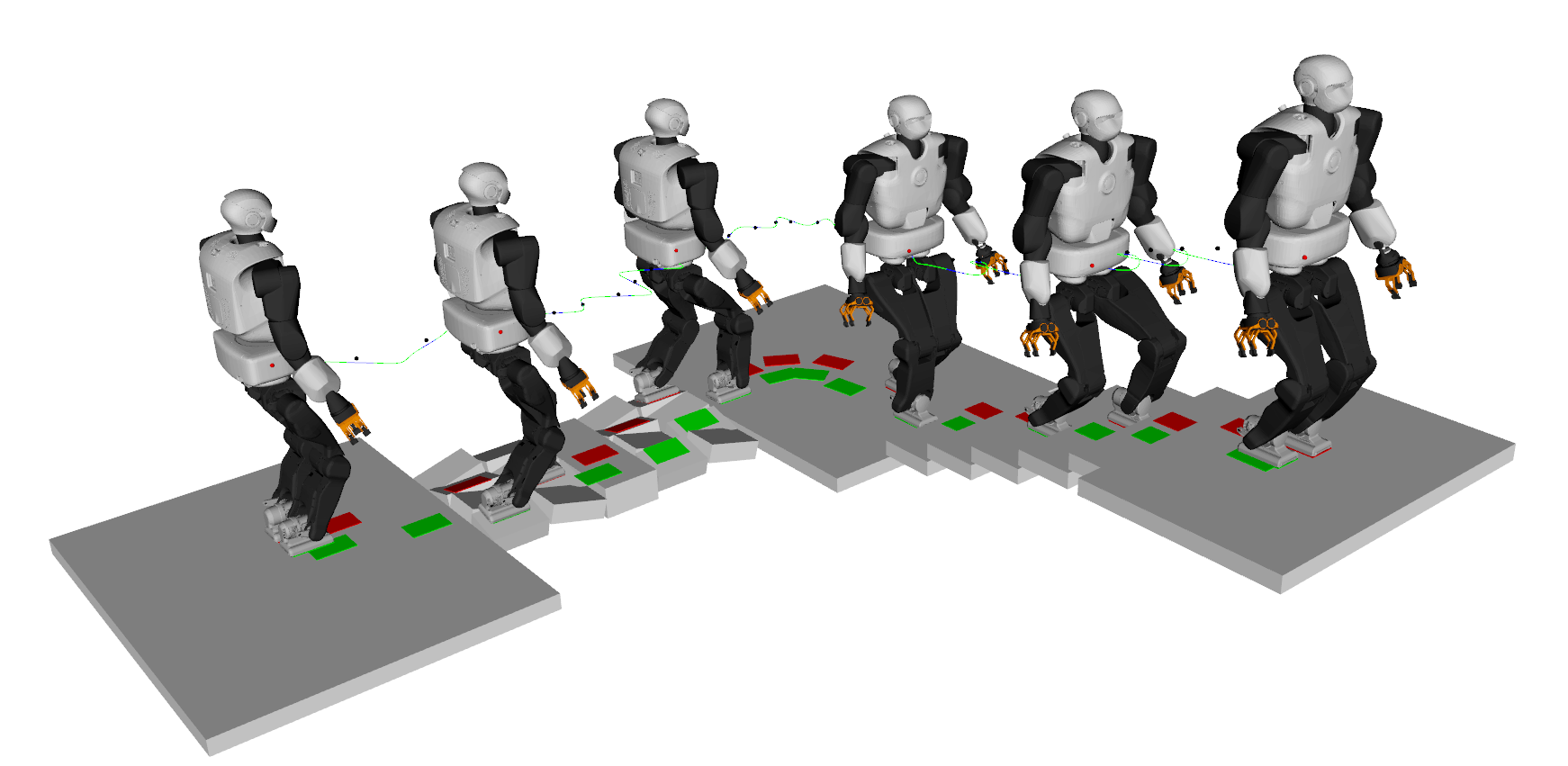}
\caption{A contact sequence generated by our convex-relaxed approach with domain-specific information. Talos robot is executing planned locomotion. } \vspace{-1.0em}
\label{fig:main}
\end{figure} 

Several relaxations have been proposed to practically address the problem of footstep planning. The model size can be reduced by approximating the robot with a point mass model~\cite{Kajita03, Wensing13, caron2019tro} and expressing the constraints as simplified functions of the Center of Mass (COM) and contact positions only~\cite{wieber,Carpentier-RSS-17, tonneau20182pac}. The price for these approximations is paid with the loss of completeness \cite{escande2013planning, fernbachCroc} and/or the guarantees of convergence \cite{Winkler, fallon2015architecture, carpentier2018multicontact}. Which part of the problem can be approximated is thus a fundamental question closely related to the issue of modelling the combinatorics. Discrete approaches embrace the discrete nature of the problem~\cite{Bretl2006MotionPO, hauser2008motion, deeploco,deepgait} , while continuous, optimisation-based approaches have to design strategies to handle the discrete variables. In \cite{mordatch2012DiscCIopt}, a continuation method is employed to first solve a simplified version of the problem. Then the solution is used as a starting point for subsequent problems of increasing difficulty until the original problem is solved. Another option is to implicitly represent the discrete variables with continuous ones and to approximate the environment as a continuous function \cite{Winkler}. Alternatively, complementarity constraints can be used to enforce contact decisions, resulting in a computationally demanding, non-smooth optimisation problem~\cite{yunt2006trajectory,posa}.

Though often conflicted in the literature, discrete and continuous approaches can be unified in the Mixed-Integer Programming (MIP) formalism~\cite{deits2014footstep, daiMIPIK, 20-driess-ICRA, ponton2020efficient, aceituno2017simultaneous}. 
Internally, MIP solvers first solve an approximated problem, in which they relax the discrete (integer) variables into continuous ones. In the best-case scenario, the relaxed variables converge to integer values, and the original problem is solved. Otherwise, the combinatorics have to be addressed with branch-and-bound strategies. In the worst case, this requires solving an optimisation problem for every possible combination of values for the integer variables. Fortunately, this is rarely needed in practice as efficient approaches exist to solve only a fraction of the combinations~\cite{gurobiprimer}. MIP solvers can efficiently handle convex problems, but struggle to address non-linear problems efficiently~\cite{wang2021automatic}, which is why MIP-based approaches such as \cite{deits2014footstep} work with linearised formulations.

In the best case, solving MIP is computationally equivalent to solving a continuous formulation of the problem, where no specific scheme is required to address the combinatorics. From this perspective, our research question is: {\em is it always possible to end up with a best-case scenario for the MIP in the context of footstep planning?} A positive answer would imply that the problem can be addressed effectively by off-the-shelf optimisation solvers, resulting in simpler and faster formulations of the contact planning problem. 
In this paper, we empirically show a positive result for a diverse range of instances of the contact planning problem, featuring stairs, rubbles, and narrow passages.

This paper is an extension of our earlier work \cite{tonneau2020sl1m}, where we proposed a convex relaxation of the MIP approach for footstep planning with an \lone-norm minimisation, SL1M. In this preliminary work \cite{tonneau2020sl1m}, SL1M has been shown to converge to integer solutions for ``simple'' problems, but fails when the combinatorial complexity becomes too high (as defined in Section~\ref{sec:sl1m}). This issue is the focus of this paper. 

\subsection{Main Contributions}
In this paper, we considerably improve our earlier work on SL1M by reducing the complexity of footstep planning problems. This is achieved by automatically removing non-relevant integer combinations using a low-dimensional path planner that computes a trajectory for the root of the robot~\cite{tonneau2018efficient}. We verified that improved SL1M always converges to an integer solution in our test scenarios after the pruning. We experimentally validate our hypothesis that we can reach a best-case scenario for the MIP in footstep planning by pruning the irrelevant combinations.

Thus, our main contribution is an LP relaxation of the MIP formulation, SL1M, experimentally shown to outperform commercial MIP solvers.
Our second contribution is a demonstration that using a trajectory planner also improves the computational performance of MIP problems, especially for complex scenarios. 
These findings are demonstrated in several scenarios involving biped and quadruped robots navigating across uneven terrains with a pre-established gait.

\section{Rationale}
We study the combinatorial aspect of contact planning that results from environmental constraints. For instance, when climbing a staircase, we must decide which steps (or contact surfaces) our foot must land on. Planning the next $n$ steps considering $m$ possible contact surfaces for each step involves solving $m^n$ optimisation problems in the worst case under the MIP formalism. We want to reformulate this problem so that we can find a solution by solving a single optimisation problem (or a few).

To achieve this objective, we first recall how to write the footstep planning problem as a MIP. Then we follow two paths of action. First, we automatically reduce the number of possible combinations. 
In our staircase example, if the robot starts at the bottom of the stairs, it is useless to consider the tenth step surface when planning the very first footstep. In general, any contact surface that is not in the reachable space of the foot 
can be discarded from the set of candidates. We automate this pruning process with a low-dimensional sampling-based trajectory planner~\cite{fernbach2017kinodynamic}.
Second, we verify that solving a feasibility problem requires fewer iterations than solving a minimisation problem. More interestingly, after recalling that the feasibility problem can be relaxed into an \lone-norm minimisation problem, we empirically show that the relaxed problem {\em always} converges to a feasible solution when the combinatorics is first reduced using our trajectory.

Conversely, our results show that currently, our continuous optimisation-based formulation does not work well if an objective has to be minimised simultaneously.
However, this issue can be alleviated: once the contact surfaces have been selected by the \lone-norm relaxation of the feasibility problem, we can solve a second problem, which minimises a cost function but without modifying the contact surfaces selected by the first problem. This second problem boils down to a simple convex Quadratic Program (QP) \cite{deits2014footstep, tonneau2020sl1m}.

\section {Definitions and Notation}

Table \ref{table:notations} defines the notation used throughout the paper. Unless specified, we use subscript for indicating the $i$-th footstep and superscript for indicating the $j$-th contact surface. 

\begin{table}[tb]
\centering
\caption{Table of Notation}
\resizebox{\linewidth}{!}{
\begin{tabular}{p{0.18\linewidth}p{0.76\linewidth}} 
 \toprule
 \textbf{Notation} & \textbf{Description} \\
 \midrule
    $n$ & the number of footsteps in the planning\\
    $m$ & the number of contact surfaces in the environment\\
    \multirow{1}{*}{$\mathcal{S}$} & union of potential contact surfaces available \\
    $\mathcal{S}^j \subset  \mathcal{S}$ & the $j$-th contact surface\\
    $\mathcal{S}_i \subset  \mathcal{S}$ & the contact surfaces considered in the $i$-th footstep\\
    $\mathbf{p}_i \in \mathbb{R}^{3}$ & $i$-th footstep position \\  
    $\mathbf{a}_{i}^{j} \in \{0,1\}$ & integer slack variable for $j$-th surface in $i$-th footstep\\
    $\bm{\alpha}_{i}^{j} \in \mathbb{R}^{+}$ & positive real slack variable for $j$-th surface in $i$-th footstep\\ 
    $\bm{\beta}_{i}^{j} \in \mathbb{R}$ & real slack variables for $j$-th surface in $i$-th footstep\\
    $\vc{R}_i \in SE(3)$ & root position / orientation when creating footstep $i$ \\
    $\mathbf{card}(\mathbf{a})$  & number of non-zero entries in vector $\mathbf{a}$ \\
    $\mathcal{I} ,\mathcal{G}$ & initial / goal constraint sets\\
    $\mathcal{F}$ & feasibility constraint set\\
    
 \bottomrule
\end{tabular}
}
\label{table:notations}
\end{table} 

\begin{figure}[tb]
\centering
\includegraphics[width=0.9\linewidth]{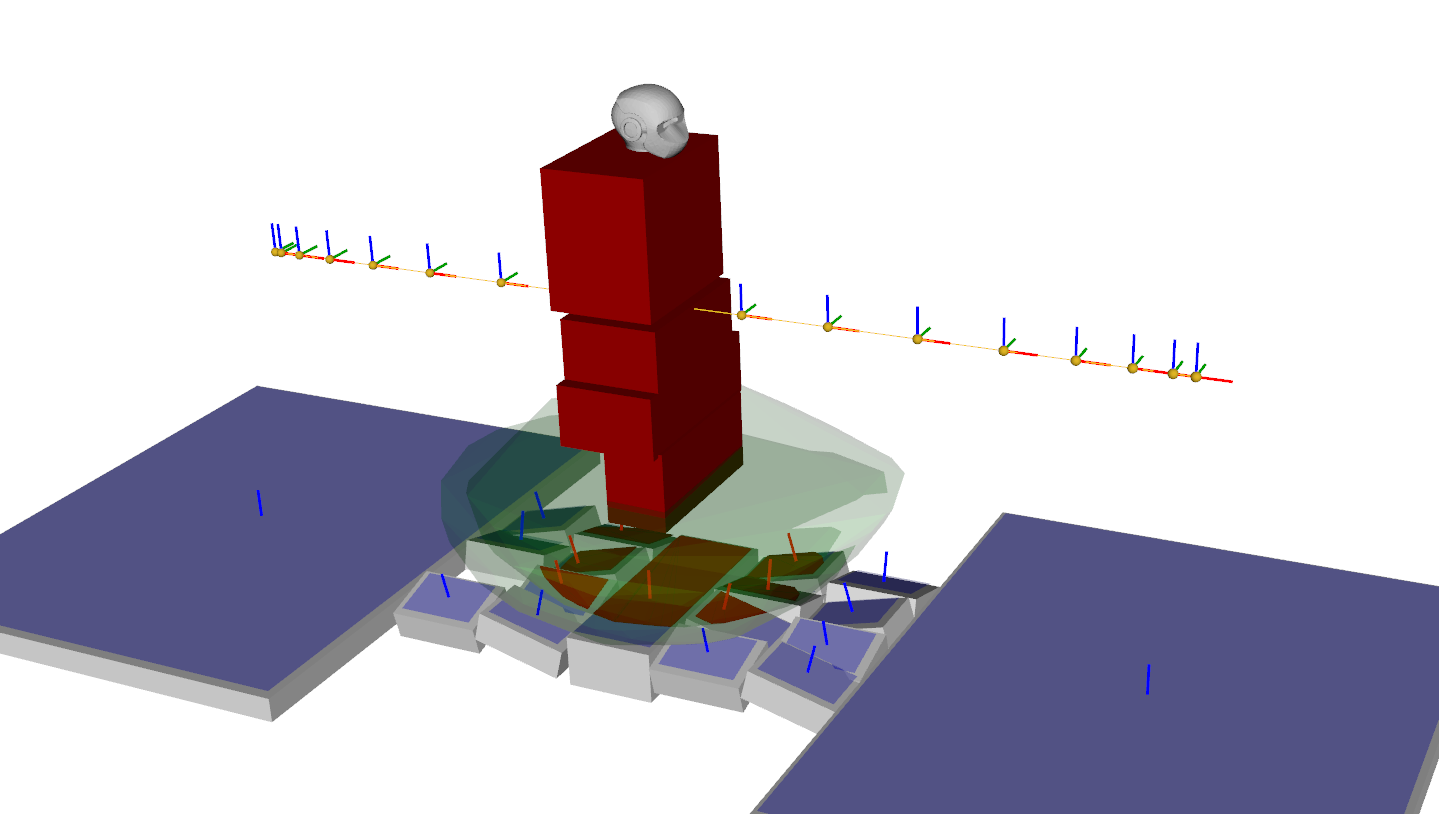}
\caption{The environment is a set of contact surfaces (blue and red). The reachable workspace of each foot is shown in green. The intersection between the reachable workspace of the left foot and the environment (red) defines potential contact surfaces. The normals of the contact surfaces (arrows) constrain the orientation of the foot around the x- and y- axes. }\vspace{-0.5em}\label{fig:intersection}
\end{figure}

We now more specifically define the environment. 
The environment is represented as a union of $m$ disjoint sets $\mathcal{S} = \bigcup_{j=1}^{m} \mathcal{S}^j$ as shown in Fig.~\ref{fig:intersection}. 
Each set $\mathcal{S}^j$ represents a convex contact surface, that is a polygon embedded in a 3D plane and bounded by a set of half-spaces:
\begin{equation}\label{eq:surf}
\begin{aligned}
    \mathcal{S}^j := \{\mathbf{p} \in \mathbb{R}^3 | \mathbf{p}^T\mathbf{n}^j = e^j,  \mathbf{S}^j \mathbf{p} \leq \mathbf{s}^j \}  \,.
\end{aligned}
\end{equation}
The equality defines the plane containing the contact surface, given by the normal  $\mathbf{n}^j \in \mathbb{R}^3$ and $e^j \in \mathbb{R}$.  $\mathbf{S}^j \in \mathbb{R}^{h\times3}$ and $\mathbf{s}^j \in \mathbb{R}^{h}$ are respectively a constant matrix and a vector defining the $h$ half-spaces that bound the quasi-flat\footnote{A surface is quasi-flat if the associated friction cone contains the gravity direction.}~\cite{delprete15} surface. $\mathcal{S}_i \subset \mathcal{S}$ is the subset of contact surfaces that are considered for the $i$-th footstep, which we call the candidate surfaces.

\section{Footstep planning as a MIP}
The section recalls casting the footstep planning problem as a MIP problem, adapted from \cite{tonneau2020sl1m} and originally introduced in \cite{deits2014footstep}.
We write the footstep planning problem as:
\begin{equation}\label{eq:footstep}
\begin{aligned}
    \textbf{find} \quad & \mathbf{P} = [\mathbf{p}_1,\cdots,\mathbf{p}_{n}] \in \mathbb{R}^{3\times n} \\
    \textbf{min} \quad & l(\mathbf{P}) \\
    \textbf{s.t.} \quad & \mathbf{P} \in \mathcal{I} \cap  \mathcal{G} \cap \mathcal{F} \\
    \quad & \mathbf{p}_i  \in \mathcal{S} \quad \forall i, 1 \leq i \leq n \,.
\end{aligned}
\end{equation}

We want to find a user-defined number (for now) $n$ of footstep positions $\mathbf{p}_{i}$. Optionally we may want to minimise an objective $l(\mathbf{P})$. $\vc{P}$ must satisfy initial and goal conditions, where simple conditions can be given by the sets $\mathcal{I} : \{\vc{P}, \mathbf{p}_{1} = \mathbf{p}_{\mathcal{I}}\}$ and $\mathcal{G} : \{\vc{P}, \mathbf{p}_{n} = \mathbf{p}_{\mathcal{G}}\}$, with user-defined constants $\mathbf{p}_{\mathcal{I}}$ and $\mathbf{p}_{\mathcal{G}}$. 
$\mathcal{F}$ denotes the set of kinematic and dynamic feasibility constraints that guarantee the robot to follow the footstep plan without falling or violating joint limits, detailed in Appendix~\ref{app:constraints}. 
We constrain the positions of $\mathbf{P}$ to lie on a surface in $\mathcal{S}$.

With the surface constraints \eqref{eq:surf}, the constraint $\mathbf{p}_i \in \mathcal{S} $ can then be formulated with the use of slack variables:
\begin{align}
    \textbf{find} \quad & \mathbf{p}_i  \in \mathbb{R}^3& \nonumber \\
    \quad & \mathbf{a}_i = [a_i^1,\cdots,a_i^{m}] \in \{0,1\}^{m}& \nonumber \\ \quad & \bm{\beta}_i = [\beta_i^1,\cdots,\beta_i^{m}] \in \mathbb{R}^{m}& \nonumber \\
    \textbf{s.t.} \quad & \mathbf{card}(\mathbf{a}_i) = m -1 \label{eq:card}&\\
    \quad & \forall j, 1 \leq j \leq m :& \\
    \label{eq:Sineq} \quad & \quad \mathbf{S}^j \mathbf{p}_i \leq \mathbf{s}^j + M a_i^j \vc{1} \\
    \label{eq:betaeq} \quad & \quad \mathbf{p}_i^T\mathbf{n}^j = e^j + \beta_i^j \\
    \label{eq:betaineq} \quad & \quad ||\beta_i^j||_1 \leq M a_i^j 
\end{align}
 with $M$ being a sufficiently large constant\footnote{This formulation is known as the ``Big M'' formulation \cite{bigm}.} and $\vc{1}$ being a vector of appropriate size filled with ones.
If $a_i^j = 0$, \eqref{eq:betaineq} implies than $\beta_i^j = 0$ and as a result \eqref{eq:surf} is satisfied and $\mathbf{p}_i \in \mathcal{S}_j$. If $a_i^j = 1$, then for a sufficiently large $M$, \eqref{eq:Sineq}, \eqref{eq:betaeq}, and \eqref{eq:betaineq} always have a solution and are effectively ignored. We define a cardinality function $\mathbf{card}(\cdot)$ that counts the number of non-zero entries in a vector. Therefore, \eqref{eq:card} ensures that one contact surface is always selected. 

The complete MIP formulation of our problem is thus:
\begin{equation}\label{eq:comp}
\begin{aligned}
    \textbf{find} \quad & \mathbf{P} = [\mathbf{p}_1,\cdots,\mathbf{p}_{n}] \in \mathbb{R}^{3\times n} \\
    \quad &  \mathbf{A} = [\vc{a}_1,\cdots,\vc{a}_n] \in \{0,1\}^{n\times m}&  \\ \quad & \bm{\beta} = [\bm{\beta}_1,\cdots,\bm{\beta}_n] \in \mathbb{R}^{n\times m}&  \\
    \textbf{min} \quad & l(\mathbf{P}) \\
    \textbf{s.t.} \quad & \mathbf{P} \in \mathcal{I} \cap  \mathcal{G} \cap \mathcal{F} \\
    \quad & \forall i, 1 \leq i \leq n :& \\ 
    \quad & \mathbf{card}(\mathbf{a}_i) = m - 1 :& \\
    \quad & \quad \forall j, 1 \leq j \leq m :& \\
    \quad & \quad \quad \mathbf{S}_j \mathbf{p}_i \leq \mathbf{s}_j + M a_i^j \vc{1} \\
    \quad & \quad \quad \mathbf{p}_i^T\mathbf{n}^j = e^j + \beta_i^j \\
    \quad & \quad \quad ||\beta_i^j||_1 \leq M a_i^j \,.
\end{aligned}
\end{equation}

\subsection*{Assumptions}
We make the following assumptions to guarantee that the MIP formulation is convex:

\begin{itemize}

\item $l (\mathbf{P})$ is a convex quadratic objective function.
\item The contact surfaces are ``quasi-flat''~\cite{delprete15} and do not intersect with one another. This limitation is due to our dynamics constraint formulation, as explained in \cite{tonneau2020sl1m}.
\item Dynamic constraints are verified by computing a ``quasi-static" trajectory for the center of mass (COM).
\item Kinematics constraints on the COM are approximated as linear inequalities. 
\item The gait (i.e. the contact order for the effectors) is given.
\item For all $\vc{p}_i$, the orientation around the $\vc{z}$ axis of the foot in contact is given. We show how the orientation can be computed automatically in Section~\ref{sec:guide-path}.
\end{itemize} 
These assumptions define a convex feasible set $\mathcal{F}$.





\section{Efficient Reduction of the Combinatorics} \label{sec:guide-path}
In \eqref{eq:comp}, for each $\vc{p}_i$ we consider the complete set $\mathcal{S}$ of potential contact surfaces. However, it is reasonable to presume that for each $\vc{p}_i$ only a subset $\mathcal{S}_i \subset \mathcal{S}$ of contact surfaces are feasible, because of the geometric and dynamic constraints that bind each footstep with its neighbors. 
We approximate the subset  $\mathcal{S}_i \subset \mathcal{S}$ by exploiting the reachability condition, as introduced in \cite{tonneau2018efficient}.

\subsection{The reachability condition}
Let us assume for now that when looking for a footstep position $\vc{p}_i$, we know the position and orientation of the root of the robot $\vc{R}_i \in SE(3)$ at the moment when the contact is created. 
In this case, a necessary condition for a surface $\mathcal{S}^j$ to be a valid candidate for $\vc{p}_i$ is that $\mathcal{S}^j$ be reachable with the foot from the root pose $\vc{R}_i$. Mathematically, we can define the discrete set of reachable surfaces for footstep $i$ as:
\begin{equation}
    \mathcal{S}_i = \{ \mathcal{S}^j \in \mathcal{S} | ROM(\vc{R}_i, F_i) \cap S^j \neq \emptyset \} 
\end{equation}
where $F_i$ is the foot of interest, and $ROM$ is the 3D range of motion of the foot given a specific root pose $\vc{R}_i$ (Fig.~\ref{fig:intersection}). 

\subsection{Kino-dynamic planning for estimating the root frames}
In \cite{fernbach2017kinodynamic}, we use the reachability condition to plan 6D trajectories for the root of the robot. Given the current frame $\vc{R}_0$ and a goal frame $\vc{R}_n$, we use a kino-dynamic RRT planner to compute a continuous root trajectory $\vc{R}(t), t \in [0,T]$ of the robot that always satisfies the reachability condition:
\begin{equation}\forall t, \forall k, 1 \leq k \leq nFeet, ROM(\vc{R}(t), F_i^k)  \cap S \neq \emptyset 
\end{equation} 
where each $F_i^k$ denotes one of the $nFeet$ feet of the robot. Additionally, the trajectory prevents a collision for whole-body motions computed in its neighbourhood in practice.

\begin{figure}[tb]
\centering
\includegraphics[width=0.9\linewidth]{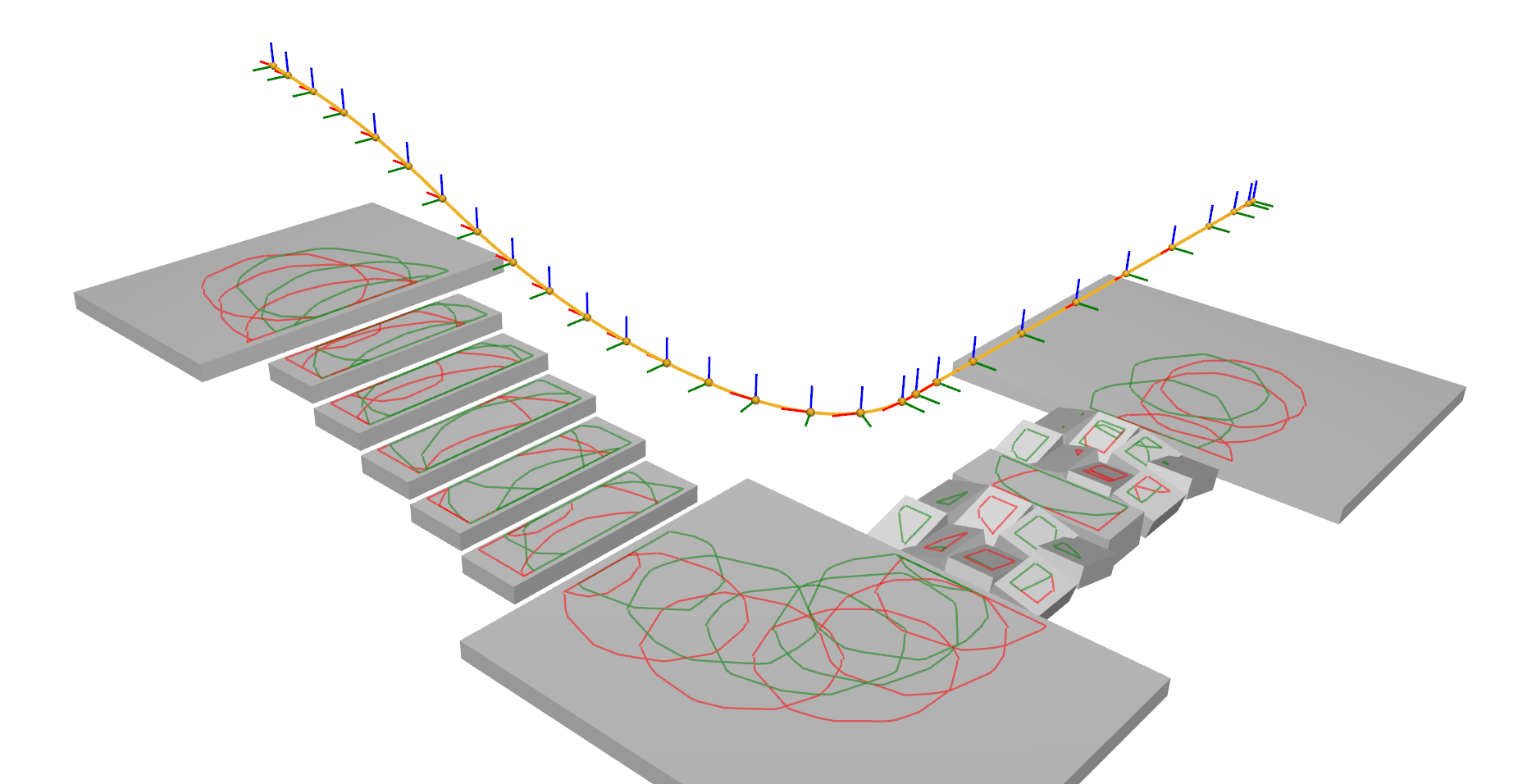}\vspace{-1.0em}
\caption{Example of a planned root trajectory (yellow) with the intersected area between the ROM and the environment (red: left foot, green: right foot) at each discrete root pose. The displayed frames show the yaw ($\vc{z}$-axis) orientation of the trajectory at the discrete points, used as the yaw orientations of the contacting feet.}\label{fig:guide-path} \vspace{-0.5em}
\end{figure}

We discretise $\vc{R}(t)$ with a time step $\delta t$, which is the only hyperparameter required in our framework. Assuming that a new step occurs every $\delta t$, we compute the number $n$ of required footsteps to reach the target, as well as the estimated root poses $\vc{R}= [ \vc{R}(\delta t), \vc{R}(2 \delta t) ,\cdots, \vc{R}(T)]$ at each step creation. 
Fig.~\ref{fig:guide-path} shows an example of an $SE(3)$ root trajectory and its discretisation. The reachable space of each foot is shown for each step. 
We assume that for each $\vc{p}_i$, the orientation of the foot about the $\vc{z}$-axis is aligned with the root. 

The kino-dynamic planner allows us to filter the contact candidates and pre-define the orientation of the feet, significantly reducing the number of surface candidates in our MIP. It is interesting to note that the information given by the kino-dynamic planner (number of steps and rotation about the $\vc{z}$-axis) is handled inside the quadratic cost in the MIP formulation by \cite{deits2014footstep}. Their formulation has the advantage of being self-contained, although it does allow for consideration of non-flat contact surfaces. The choice of optimising the number of steps and orientation within the cost, as opposed to our potentially sub-optimal approach, deserves a discussion. We believe that the tuning required to weigh the several terms in the cost is as hard as the parametrisation of the $\delta t$ parameter. The potential sub-optimality is the price to pay for the computational gains shown in our results, which are connected to the reduced combinatorics by our approach. We detail the kino-dynamic root trajectory planner in Appendix~\ref{app:planner}.

\section{Convex Relaxation of the MIP feasibility problem} \label{sec:sl1m}
We purposely formulated \eqref{eq:comp} as a special instance of a \emph{cardinality problem}~\cite{vandenberghe1996semidefinite,lemarechal1999semidefinite}. 
We observe that the constraint $\mathbf{card}(\mathbf{a}_i) = m - 1$ constrains $\mathbf{card}(\mathbf{a}_i)$ to its minimum: as all the $\mathcal{S}^j$ are disjoint, it is impossible for a footstep to lie simultaneously on more than one contact surface. Therefore, by removing the cardinality constraint and replacing the cost with  $l(\vc{P}) + w \sum_{i=1}^n \mathbf{card}(\vc{a}_i)$, with $w$ being a sufficiently large weight, we obtain a strictly equivalent problem.

Cardinality minimisation problems are well-known to be efficiently approximated with \lone-norm minimisation problems, thanks to the sparsity induced by the \lone-norm~\cite{boydCardinalsity,abdi2013cardinality}. Therefore, we replace the Boolean variables $\vc{a}_i$ with the real variables $\bm{\alpha}_i$ and minimise their norm to obtain what we call SL1M formulation~\cite{tonneau2020sl1m}. Combining the \lone-norm relaxation with the pruning obtained through the use of the kino-dynamic planner as proposed in Section \ref{sec:guide-path}, we obtain the following problem:
\begin{equation}\label{eq:sl1m_complete}
\begin{aligned}
    \textbf{find} \quad & \mathbf{P} = [\mathbf{p}_1,\cdots,\mathbf{p}_{n}] \in \mathbb{R}^{3\times n} \\
    \quad &  \bm{\alpha} = [\bm{\alpha}_1,\cdots,\bm{\alpha}_n], \bm{\alpha}_i \in \mathbb{R}^{+ |S_i|}&  \\ \quad & \bm{\beta} = [\bm{\beta}_1,\cdots,\bm{\beta}_n], \bm{\beta}_i \in \mathbb{R}^{|S_i|}&  \\
    \textbf{min} \quad & l(\mathbf{P}) + w \sum_{i=1}^n \sum_{j=1}^{|S_i|} \alpha_i^j \\
    \textbf{s.t.} \quad & \mathbf{P} \in \mathcal{I} \cap  \mathcal{G} \cap \mathcal{F} \\
    \quad & \forall i, 1 \leq i \leq n :& \\ 
    \quad & \quad \forall j, 1 \leq j \leq |\mathcal{S}_i| :& \\
    \quad & \quad \quad \mathbf{S}_{i,j} \mathbf{p}_i \leq \mathbf{s}_{i,j} + M \alpha_i^j \vc{1} \\
    \quad & \quad \quad \mathbf{p}_i^T\mathbf{n}_{i,j} = e_{i,j} + \beta_i^j \\
    \quad & \quad \quad ||\beta_i^j||_1 \leq M \alpha_i^j 
\end{aligned}
\end{equation}
where $|\mathcal{S}_i|$ is the number of elements of the set $\mathcal{S}_i$. $\mathbf{S}_{i,j}$, $\mathbf{s}_{i,j}$, $\mathbf{n}_{i,j}$ and $e_{i,j}$ are the constants defined for the $j$-th surface inside the set $\mathcal{S}_{i}$. We observe that since all the elements of $\bm{\alpha}$ are positive, $||\bm{\alpha}_i^j||_1 = \bm{\alpha}_i^j, \forall i,j$. We set $l(\mathbf{P})=0$ and $w = 1$ when solving a feasibility problem. This transforms the optimisation problem into a feasibility problem. Once the contact surfaces are fixed, the quadratic cost $l(\mathbf{P})$ is reintroduced to locally optimise the footstep positions on the selected surfaces.

For a solution of \eqref{eq:sl1m_complete} to be a solution of \eqref{eq:comp}, we need that $\mathbf{card}(\bm{\alpha}_i) = m-1, \forall i$. However, this might not be the case. To handle any potential non-sparse optimum, we use a brute-force approach. We fix all the variables $\bm{\alpha}_i$ for which the cardinality constraint is satisfied. We then test all the combinations for the remaining free variables until either \rom{1} a solution is found, \rom{2} a maximum number of trials is reached, or \rom{3} all possibilities are exhausted. This approach is not fail-proof, but this is a good compromise for our goal of finding a solution by solving a minimum number of optimisation problems. 

An interesting observation is that because we are solving a cardinality problem, \eqref{eq:sl1m_complete} is very close to the relaxed problem initially solved by a MIP solver for \eqref{eq:comp}. It would be equivalent if the cardinality constraint was also reformulated as part of the cost and the pruning of contact surfaces was included in \eqref{eq:comp}. In Section \ref{sec:result}, we will show that because of the preliminary operations run by a MIP solver, it is more computationally efficient to directly solve \eqref{eq:sl1m_complete} if the problem converges to a feasible solution.

\subsection {Extension to quadruped locomotion}
In terms of discrete variables, the proposed approach extends directly to any number of legs as long as the gait is pre-defined, leaving \eqref{eq:sl1m_complete} and \eqref{eq:comp} unchanged.
A conservative change must however be brought into the quasi-static constraints defining $\mathcal{F}$ to guarantee a convex formulation. Quasi-static bipedal locomotion requires the COM to lie above the foot when the other foot breaks contact. For a quadruped, this constraint is not as limiting, as the COM can lie anywhere in the convex hull of the current contact points. However, if both the COM and feet locations are variables, the convex hull constraint is not convex. To preserve convexity, we express the COM's x and y coordinates as linear functions of the positions of the effectors (i.e. feet) in contact:
\begin{equation}\label{eq:equi_multi_contact}
\begin{aligned}
    \mathbf{c}_{i_{x,y}} = \sum_k w^k_i \mathbf{p}^k_{i_{x,y}}, \quad \sum_k w^k_i = 1, \quad w^k_i \ge 0 \quad \forall i 
\end{aligned}
\end{equation}
where $\mathbf{p}^k_i$ represents the position of the $k$-th effectors in $i$-th contact, and $w^k_i$ is a user-defined unit weighting vector that is fixed for each contact. In our tests, we average the weights, writing $w_i^k = 1/size(w_i)$.

\begin{table*}[ht] \vspace{1.0em}
\centering
\caption{Contact Planning Computation Performance Evaluation}
\begin{tabular}{rcccccccccc} 
 \toprule
 \multirow{2}{*}{\textbf{Scenario}} & \multirow{2}{*}{\textbf{\# footstep}} & \multicolumn{4}{c}{\textbf{w/o trajectory planning}} & \multicolumn{5}{c}{\textbf{w/ trajectory planning}} \\
 \cmidrule(lr{0.5em}){3-6} \cmidrule(lr{0.5em}){7-11} 
                                    &  & \# surf. & MIP opt. & MIP feas. & SL1M & avg. \# surf. & trajectory & MIP opt. & MIP feas. & SL1M \\
 \midrule
    bridge & 16 &  3 & 157.5     & 166.2     &  45.3     & 1.2 & 179.4     & 87.5     & 77.3     & 35.5     \\  %
    stairs &  12 &  7 & 130.3     & 59.5     & 23.6     & 2.0 & 319.0     & 92.6     & 34.7     & 19.9     \\ 
    \hline
    rubbles & 16 & 18 & 1914.0     & 389.4     & - & 5.9 & 252.2     & 647.7     & 230.2     & 73.4     \\ 
    rubbles \& stairs & 32 & 24 &98060.5     &  \cellcolor{yellow}74710.5     & - & 3.9 & \cellcolor{yellow}501.8     & 1070.5     & 508.9     &  \cellcolor{yellow}217.6     \\ 
 \bottomrule
\end{tabular}
\label{table:comptime}
\end{table*} 

\begin{figure*}[ht] 
\centering
\includegraphics[width=0.45\linewidth]{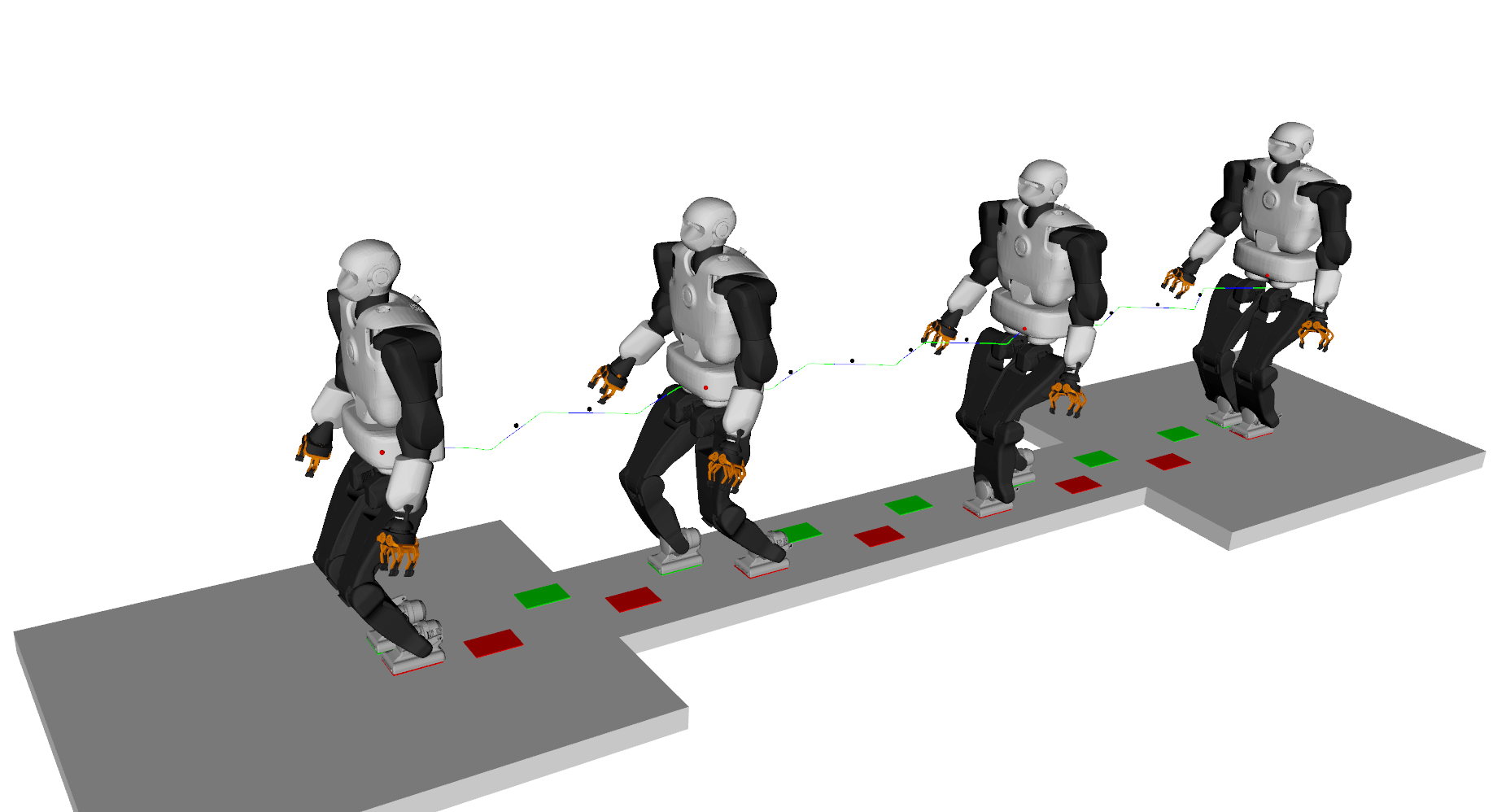}\hspace{-0.3em}
\includegraphics[width=0.45\linewidth]{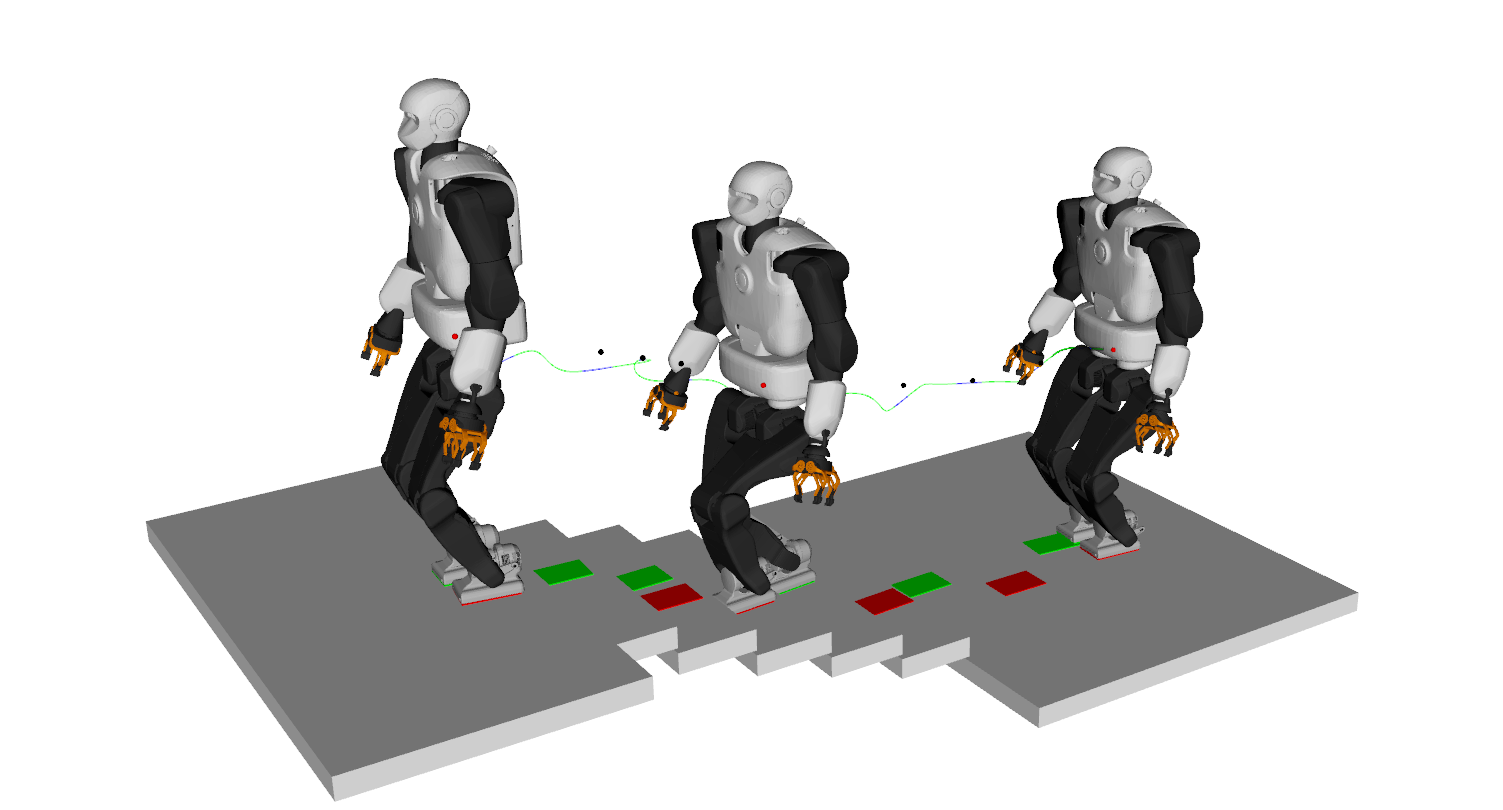}
\includegraphics[width=0.45\linewidth]{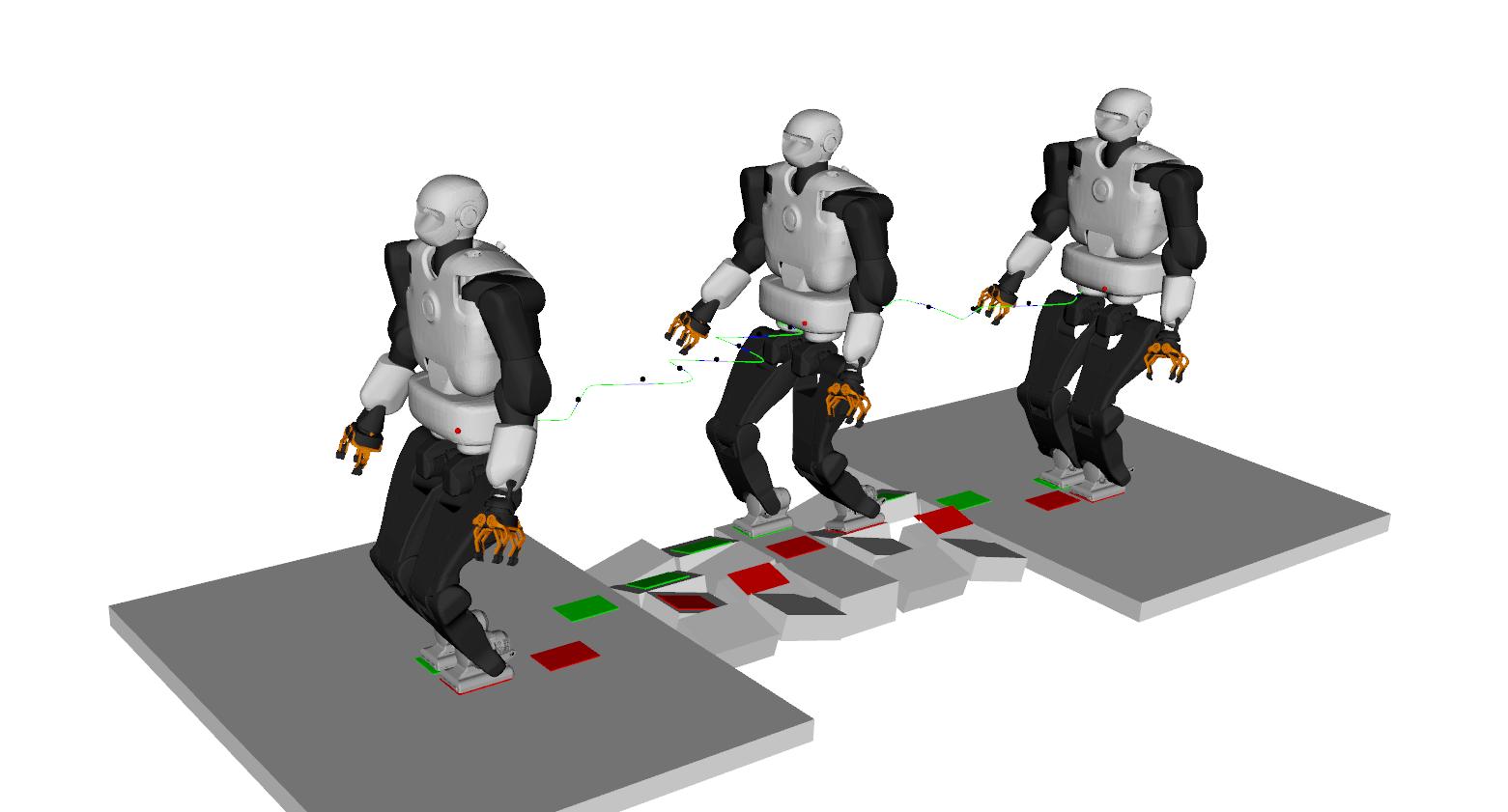}\hspace{-0.3em}
\includegraphics[width=0.45\linewidth]{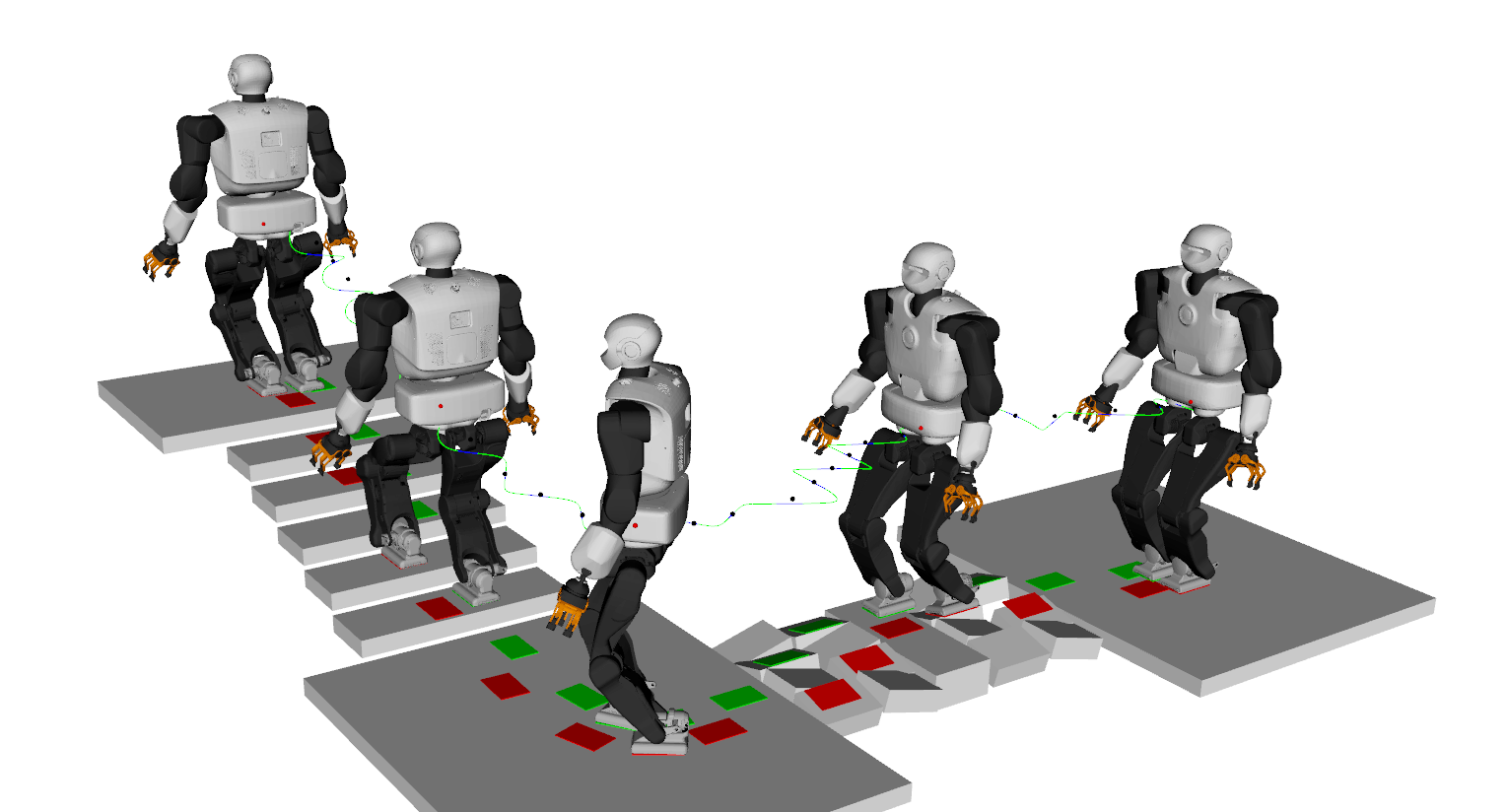} 
\caption{Results of contact planning and the whole-body motion planning of Talos in our sample scenarios. The top row shows the \textit{easy problems}, bridge and stairs. The bottom row shows the \textit{challenging problems}, rubbles and rubbles \& stairs.} \vspace{-0.5em}\label{fig:scenes} 
\end{figure*} 

\section{Experimental Results} \label{sec:result}
We tested our approach in simulation in a variety of environments for the humanoid Talos~\cite{stasse2017talos} and the quadruped ANYmal~\cite{hutter2016anymal}.
To experimentally validate our framework, we provide both quantitative and qualitative comparisons against the different approaches explained. 
These results highlight the performance improvements by reducing the combinatorial complexity of the footstep planning problem using a trajectory. 

\subsection{Implementation details} 
The test environment is written in Python. The actual initialisation and resolution of both \eqref{eq:comp} and \eqref{eq:sl1m_complete} are written in C++ and invoked through dedicated Python bindings. 
All optimisation problems are solved using the Gurobi \cite{gurobi} solver through those bindings\footnote{The informed reader will observe that Python bindings already exist for Gurobi. However, the initialisation of a problem through Python is not computationally efficient, which justified the writing of a dedicated C++ library. This inefficiency also explains the difference in the computation times observed for the MIP resolution in the current work and in \cite{tonneau2020sl1m}.}. 
Open-source LP/QP solvers such as Quadprog \cite{quadprog} and GLPK \cite{makhorin2008glpk} give similar performance results for the case of SL1M. However, this is not true for the MIP formulation. We chose Gurobi for a fair comparison. The measured computation times represent the time spent in the initialisation and resolution of the optimisation problem by the solver. 
The trajectory planner is implemented in C++ within the HPP framework~\cite{mirabel2016hpp} and invoked from Python using a CORBA architecture. Tests were run on a PC with an AMD Ryzen 7 1700X eight-core processor on Ubuntu 18.04. 

\subsection{Quantitative analysis}

We selected four environments representative of the difficulty of a combinatorial footstep planning problem (Fig.~\ref{fig:scenes}). In each scenario, the objective is to find a sequence of footsteps resulting in a feasible whole-body motion containing the given initial and goal root poses. 
For each scenario, we solve the footstep planning problem using three methods: MIP optimisation, MIP feasibility, and SL1M. 
MIP optimisation corresponds to \eqref{eq:comp}, where we set an objective function, just for comparison, that minimises the sum of the squared distances between the planned footsteps.
MIP feasibility and SL1M represent \eqref{eq:comp} and \eqref{eq:sl1m_complete} respectively with $l(\mathbf{P}) = 0$, solving a feasibility problem. 
Once the contact surfaces are fixed, we locally optimise the footstep positions on the selected surfaces. To achieve this, we call an instance of \eqref{eq:comp}, where all integer variables are fixed and the quadratic cost $l(\mathbf{P})$ is reintroduced.

In addition, each method is tested twice with the full combinatorics and with reduced combinatorics based on the trajectory. 
The same number of footsteps $n$ computed by discretising the trajectory is used in both cases.

Table \ref{table:comptime} reports the computation times in milliseconds of each method averaged over 100 runs, together with the number of footsteps and the average number of candidate surfaces per contact. Empty cells mean that the method was not able to converge to a feasible solution. In all other cases, the success rate was 100\% for all the methods.

\subsubsection{Easy scenarios (bridge and stairs)}
First, we consider the environments composed of less than 10 candidate surfaces. The results show that integrating the trajectory always reduces the computation times of the optimisation problem, but the total computation time (including the guide path computation) is higher. SL1M always outperforms the other methods, being up to 7.1 times faster than MIP optimisation and up to 3.7 times faster than MIP feasibility. 

\subsubsection{Hard scenarios (rubbles and rubbles \& stairs)}
We also consider environments composed of more than 10 candidate surfaces, including sloped surfaces. 
We decided to mark SL1M without trajectory planning as a failure (blank in Table \ref{table:comptime}) as the remaining combinatorics required more than 4,000 trials (Section \ref{sec:sl1m}). 
Table \ref{table:comptime} shows that SL1M with trajectory planning configuration always converges and is also always the most computationally efficient.

The pruning improved the performance by a factor of 146.8 in the best case in the stairs \& rubbles scenario with the MIP feasibility formulation. SL1M always outperforms the other methods and is up to 8.8 times faster than MIP optimisation and up to 3.1 times faster than MIP feasibility. 

Table \ref{table:opt} reports the optimal values obtained by both MIP and SL1M using a cost function that minimises the sum of squared distance traveled by the feet at each step. The MIP formulation always converges to the global optimum and thus provides a ground truth. SL1M first computes a feasible set of contact surfaces and then optimises the positions in a later stage. As expected, when MIP and SL1M select the same contact surfaces, the optimal cost is the same (bridge). Conversely, the cost differs when the contact surfaces selected differ, which is expected as SL1M is only computing a feasible solution. We also observe that the pruning based on the kino-dynamic planner always precludes the solver from finding the global optimum. However, we observe that the increase in the resulting cost is small (always less than 10\%), which we consider to be an acceptable trade-off.

\begin{table}[tb] 
\centering
\caption{Cost Function Values}
\resizebox{\linewidth}{!}{
\begin{tabular}{rcccc} 
 \toprule
 \multirow{2}{*}{\textbf{Scenario}} & \multicolumn{2}{c}{\textbf{w/o trajectory}} & \multicolumn{2}{c}{\textbf{w/ trajectory}} \\
 \cmidrule(lr{0.5em}){2-3} \cmidrule(lr{0.5em}){4-5} 
                                   & MIP opt. & SL1M + QP & MIP opt. & SL1M + QP\\
 \midrule
    bridge &  1.887 & 1.887  & 1.974 & 1.974  \\ 
    stairs & 0.797 & 0.872   & 0.870  & 0.929 \\ 
    rubbles &  0.968 & -  & 1.022 & 1.071  \\ 
    rubbles \& stairs &  1.880 & -  & 2.072 & 2.224   \\ 
    
 \bottomrule
\end{tabular}
}\vspace{-0.5em}
\label{table:opt}
\end{table} 


\subsection{Qualitative validation}
\begin{figure}[tb]
\centering
\includegraphics[width=0.9\linewidth]{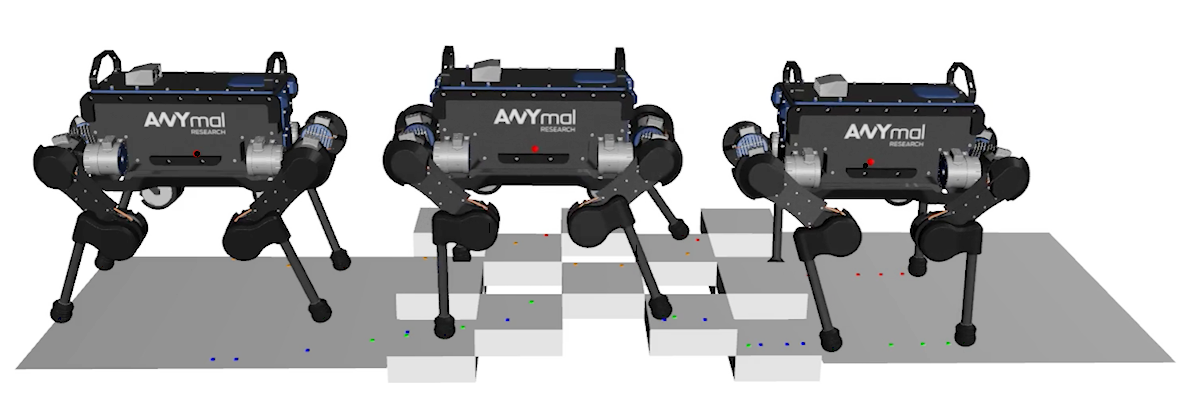}
\vspace{-0.5em}\caption{Contact planning results for ANYmal crossing the palet. The colored cubes indicate the planned contact for each foot.} 
\label{fig:quadruped}
\end{figure}

We performed whole-body motion generation using \cite{fernbachCroc,adelprete:jnrh:2016} to validate the computed footstep plans. Fig.~\ref{fig:scenes} shows the snapshots of the generated motions for Talos along with the planned contact sequences. Although there are no theoretical guarantees that the contact plans can be extended to feasible whole-body motions (beside the quasi-static guarantee), this was the case in all of our scenarios. The companion video shows the resulting motions and compares them with the MIP formulation. The contact surfaces are selected using SL1M with trajectory planning, and then the footstep positions are optimised with a QP formulation. We also present a qualitative result obtained with the quadruped (Fig.~\ref{fig:quadruped}). 

\vspace{0.5em}
\subsection{Complete Pipeline}

Table \ref{table:complete} shows the average computation time in milliseconds, profiling the pipeline in SL1M with trajectory planner configuration. The computation times are broken down into \rom{1} trajectory planning, \rom{2} contact surface selection with SL1M, and \rom{3} QP-based footstep location optimisation. 

\begin{table}[b] 
\centering
\vspace{-0.5em}\caption{Computation Time for the Complete Pipeline}
\resizebox{\linewidth}{!}{
\begin{tabular}{rcccc} 
 \toprule
 \textbf{Scenario} & \textbf{Trajectory} & \textbf{Footstep} & \textbf{Optimisation} & \textbf{Total}\\
 \midrule
    bridge &  179.4    & 35.5    & 6.9    & 221.8   \\  
    stairs &  319.0    & 19.9    & 7.5    & 346.4    \\ 
    rubbles & 252.2    & 73.4    & 8.0    & 333.6   \\ 
    rubbles \& stairs & 501.8    & 217.6    & 19.3    & 738.7   \\ 
 \bottomrule
\end{tabular}
}
\label{table:complete}
\end{table} 

\vspace{0.5em}
\section{Discussion} \label{sec:discussion}

From these experiments, we draw two conclusions. First, decreasing the possible combinations with a trajectory planner reduces the computation time needed to solve the tested optimisation problems. The improvement ranged from over 146.8 times to 1.180 times. It also improved the convergence of SL1M in challenging scenarios. Pruning the non-relevant candidates was more effective in challenging scenarios than in easy scenarios with a stronger reduction on the complexity (${\# surf}^{\# footstep}$) of the problem. Indeed in easy scenarios, the time required for pruning outweighed the time savings in the optimisation phase. Second, solving feasibility problems (SL1M, MIP feasibility) is faster than solving optimisation problems (MIP optimisation) in general. 

\paragraph* {Always use the trajectory planner}
Using a trajectory planner is not efficient for \textit{easy problems} in terms of computation time. However, it brings other advantages such as automatic constraint construction and foot orientation information. Therefore, we recommend always using it. For \textit{hard problems}, computational gains are remarkable. For instance, in the rubbles \& stairs scenario, it is more than 100 times faster when compared with MIP feasibility without trajectory planning (yellow mark in Table~\ref{table:comptime}).

\paragraph* {Non-sparse optimum handling}
The \lone-norm relaxation can lead to a non-sparse optimum solution, where the relaxed integer variables do not converge to an integer value. A brute-force approach to handle this non-sparse optimum normally works well with easy problems. However, with challenging problems, we saw that it can result in large combinatorics. We tested specific methods to enhance the sparsity of \lone-norm minimisation, such as iterative re-weighting \cite{candes2008enhancing}. This did not improve the success rate for our problem, while in comparison, trajectory pruning was proved to be more efficient.

\paragraph*{Combinatorial reduction in MIP problems}
Before solving a combinatorial problem, MIP solvers commonly perform a ``\emph{presolve}'' \cite{gurobiprimer}, where the problem is analysed, and unfeasible solutions are removed from the search space. Our trajectory planner can be considered as a domain-specific implementation of the presolve routine.
Studying the influence of the presolve in challenging scenarios gives us some relevant insight into their combinatorics. The explored branch-and-bound node counts in MIP are shown in Table \ref{table:presolve}. 0 means that MIP converged to integer values before getting into a branch-and-bound phase. 
When the trajectory is combined and the presolve is enabled, we can see that the MIP feasibility problems directly converge to a feasible solution. This is expected, as the first iteration of the MIP is equivalent to SL1M. More interestingly, this is also the case for the MIP optimisation problems, where 2960 nodes had to be explored to find the optimum without the trajectory. These results suggest that for our scenarios, even when considering a cost function, \emph{there is no combinatorics involved in the resolution of the footstep planning problem}. Studying the validity of this hypothesis is an exciting direction for future work.

\begin{table}[tb] 
\centering
\caption{MIP Explored Node Counts}
\resizebox{\linewidth}{!}{
\begin{tabular}{@{}rccccc} 
 \toprule
 \multirow{2}{*}{\textbf{Scenario}} & \multirow{2}{*}{\textbf{Presolve}} & \multicolumn{2}{c}{\textbf{w/o trajectory}} & \multicolumn{2}{c}{\textbf{w/ trajectory}} \\
 \cmidrule(lr{0.5em}){3-4} \cmidrule(lr{0.5em}){5-6} 
                                   & &  MIP opt. & MIP feas. & MIP opt. & MIP feas. \\
 \midrule

    rubbles & disabled & 2742 & 383  & 117 & 1  \\ 
    rubbles \& stairs & disabled & 20542 & 2724   & 7748  & 1 \\ 
    \hline
    rubbles & enabled & 1 & 1  & 0 & 0  \\ 
    rubbles \& stairs & enabled & 2960 & 1  & 0 & 0   \\ 
    
 \bottomrule
\end{tabular}
}\vspace{-1.0em}
\label{table:presolve}
\end{table} 

\paragraph*{Parameter tuning}
Our framework requires the tuning of the discretisation step $\delta t$ for the trajectory, as $\delta t$ is used to infer the number of footsteps required by the motion (a new footstep is created every $\delta t$). 
If $\delta t$ is too large, the problem can become unfeasible (not enough footsteps), while a small value increases the complexity of the problem (more footsteps and variables) and potentially leads to a failure in the convergence of SL1M. 
We heuristically selected a value $\delta t=1.0 s$ for our scenarios.
We believe $\delta t$ plays a central role in the performance of the framework and will investigate on automatically determining its value in future work.

\paragraph* {SL1M with a cost function}
Using SL1M jointly with an additional objective as in \eqref{eq:sl1m_complete} is challenging. With the current state of our knowledge, we do not recommend this approach as it increases the risk of not converging to an integer solution. However, our results show that optimising the contact locations \textit{after} selecting the contact surfaces leads to empirically convincing results, with a maximum increase of 10\% of the optimal cost. Furthermore, considering the simplified models in footstep planning, the notion of optimality is loosely related to the optimality of the resulting whole-body motion. Therefore, the suboptimality may not be critical in practice.

\section{Conclusion} \label{sec:conclusion}

We presented a convex optimisation framework for planning the footsteps of a legged robot walking on uneven terrain. Our results suggest a positive answer to the question: \textit{Can we efficiently address the combinatorics of the footstep planning problems with continuous optimisation methods?} 

We showed that in our use-cases, the footstep planning problem can be relaxed as an \lone-norm minimisation problem (SL1M), converging to a feasible solution when the combinatorics involves less than 10 surface candidates per footstep. Problems with larger combinatorics can be pruned thanks to a kino-dynamic planner, which outputs an approximation of the trajectory followed by the root of the robot.

The benefits of the framework were measured in terms of the computational gains (more than 100 times faster than the original Mixed-Integer Program in the best case), as well as the simplicity of the approach, which can be implemented using off-the-shelf open-source numerical solvers at the cost of losing guarantees of optimality. Our analysis of the behaviour of MIP solvers 
further suggests that the combination of the kino-dynamic planner and pruning methods from the literature can potentially remove the combinatorics of the problem, even when an optimal solution is desired. 





%

\appendices
\section{Reachability Constraints} \label{app:constraints}
We detail our kinematic and dynamic (here, static equilibrium) feasibility constraints, noted as a set $\mathcal{F}$ in our formulation. Two sets of constraints guarantee that the robot can follow the footstep plan without falling or violating joint limits. First, the position of the COM is constrained with respect to the contact points, following the 2-PAC approach~\cite{tonneau20182pac}. This allows to continuously guarantee feasibility of the COM trajectory while only considering two COM positions at each contact phase. We recall the method for completeness and extend it to handle variable foot translations under the quasi-flat constraint.
Second, the position of each effector is constrained with respect to the other effector in contact. This appendix is largely a reproduction of the constraints expressed in the original SL1M paper \cite{tonneau2020sl1m}.

\vspace{-0.5em}
\subsection{Center Of Mass constraints}

To guarantee that a dynamically feasible trajectory exists for the $i$-th footstep, we use the 2-PAC formulation. We only need to choose 2 COM positions for each footstep, namely $\mathbf{c}_{i,0}$ and $\mathbf{c}_{i,1}$, to guarantee continuous feasibility. Let us define a \emph{phase} as the time window between the moments two consecutive contacts are made. 

\subsubsection{Equilibrium constraints} 

For quasi-flat contact surfaces, a sufficient condition for the COM to allow for static equilibrium is: \mbox{$\cv_i \in \mathbf{conv}_i$}~\cite{delprete15}, where $\mathbf{conv}_i$ denotes the convex hull of all the contact points at phase $i$. 
For bipedal walking, this boils down to having the COM on top of the support foot.
In this case $\mathbf{c}_{i,0}$ is constrained to lie above the support polygon of $\mathbf{p}_{i-1}$ (i.e., the support foot used in the transition from phase $i-1$ to $i$, 
which was the swing foot for phase $i-1$) at the beginning of phase $i$. We then constrain $\mathbf{c}_{i,1}$ to be above $\mathbf{p}_{i}$ at the end of phase $i$ :
\begin{equation}\label{eq:equi_f}
\begin{aligned}
\mathbf{F}_{i-1}^{j} (\mathbf{c}_{i,0} - \mathbf{p}_{i-1} ) & \le \mathbf{f}_{i-1}^{j} + \mathbf{1} \alpha_{i-1}^{j} \\
\mathbf{F}_i^{j}(\mathbf{c}_{i,1} - \mathbf{p}_{i} )       & \le \mathbf{f}_i^{j}  + \mathbf{1} \alpha_i^j
\end{aligned}
\end{equation}
where $\mathbf{F}_i^{j}$ and $\mathbf{f}_i^{j}$ are the matrix and vector defining the polygonal shape of the foot associated to phase $i$ on surface $\mathcal{S}^j$.
Note that these constraints depend only on the xy coordinates of the COM and the foot positions.

By convexity of the static equilibrium region, all points of the straight line segment $[{c}_{i,0},{c}_{i,1}]$ satisfy the static equilibrium constraint. 
Similarly, the straight line segments $[{c}_{i-1,1}, {c}_{i,0}]$ and $[{c}_{i,1}, {c}_{i+1,0}]$ are also feasible because the COM stays above the support effector for all the duration of the single support phase.

\subsubsection{Reachability constraints}
We additionally constrain  $\mathbf{c}_{i,0}$ and $\mathbf{c}_{i,1}$ to guarantee kinematic reachability. 
We stress that the kinematic constraints are only approximated here, thus the ``guarantees'' that we mention for feasibility are only valid for this simplified representation of the robot.
The COM positions are linearly constrained as follows.
First, for each effector we compute offline a polytope that approximates the reachable COM workspace: a large number of configurations of the robot are randomly sampled,
and those who are collision-free and correspond to a ``quasi-flat'' pair of contacts are kept. For each of those configurations, the COM is expressed in the frame
of a given effector. The convex hull of all the computed COM positions approximates the COM workspace in the effector frame.
For each effector $k$, we thus obtain a 3D polytope $\leftidx{^k}{\mathcal{R}} : \{ \cv \in \mathbb{R}^{3}, \leftidx{^k}{\vc{R}} \cv \leq \leftidx{^k}{\vc{r}} \} $. 

At contact phase $i$, for each contact surface $\mathcal{S}^j$ the orientation of the foot frame is constant. The yaw is given by the trajectory planner, while the roll and pitch are given by the surface orientation. 
We note $\mathcal{R}{_i^{j}}$ the rotated polytope associated with contact $\mathbf{p}_{i}$ at phase $i$, assuming 
it lies on surface $\mathcal{S}^j$. The translation is variable, thus the constraints depend linearly on the effector positions. Both COM positions $\cv_{i,m}, m \in \{0,1\}$ at phase $i$ are thus constrained by the two active contacts $\mathbf{p}_{i}$ and $\mathbf{p}_{i-1}$:
\begin{equation}\label{eq:kin_com}
\begin{aligned}
\vc{R}{^j_l} (\cv_{i,m} - \mathbf{p}_{l}) \leq \vc{r}{^j_l} + \mathbf{1} \alpha_l^j && \forall j, \forall l \in \{i-1,i\} \,.
\end{aligned}
\end{equation}

Here again, the slack variable $\boldsymbol\alpha$ is used such that only the constraints related to the selected contact surfaces are applied. By convexity of our (approximated) kinematic constraints, if they are satisfied for $\mathbf{c}_{i,0}$ and $\mathbf{c}_{i,1}$ for all $i$ then they are continuously satisfied.

\subsection{Relative foot position constraints}

Similarly to the case of the COM reachability, we use a sampling-based approach to approximate the reachable workspace of each foot with respect to the others.
For effector $k$, we obtain a polytope $\leftidx{^k}{\mathcal{Q}} : \{ \mathbf{p} \in \mathbb{R}^{3}, \leftidx{^k}{\vc{Q}} \mathbf{p} \leq \leftidx{^k}{\vc{q}} \} $ that constrains the other effector. If $k$ is the moving effector at phase $i$ on surface $\mathcal{S}^j$, we write the associated constraint set $ \leftidx{^k}{\mathcal{Q}_i}^j$.
We then apply the same reasoning as for the COM to obtain the following constraints at each phase (omitting the $k$ for clarity):
\begin{equation}\label{eq:kin_rel}
\begin{aligned}
\vc{Q}{_{i-1}^j} (\mathbf{p}_i - \mathbf{p}_{i-1}) \leq \vc{q}{_{i-1}^j} + \mathbf{1} \alpha^j_{i-1} &&  \forall j \,.
\end{aligned}
\end{equation}

\section{Kino-dynamic Root Trajectory Planner}\label{app:planner}

We detail the kino-dynamic root trajectory planner \cite{fernbach2017kinodynamic} used to efficiently reduce the combinatorics as proposed in Section \ref{sec:guide-path}. 
The planner is implemented as a standard RRT algorithm with a specific steering method\footnote{In a RRT planner, a graph of states is built by sampling random states and using a local 'steering method' to try to connect them. Linear interpolation is the most trivial option. Once the local trajectory has been computed, a validator checks whether the local trajectory satisfies the constraints of the system. If part of the trajectory satisfies the constraints, the last valid state is added to the graph and the partial trajectory is kept as the edge connecting the two states} to guarantee that every position of the root in the trajectory satisfies the reachability condition. Additionally the dynamic constraints of the system are approximated with a heuristic in order to obtain smoother trajectories for the root, empirically shown to be more easily extended to feasible motions.

Concretely the steering method augments the well-known Double Integrator Minimum Time (DIMT) method~\cite{Krger2006TowardsOT} to compute minimum time trajectories between two states of the robot considering the dynamic constraints applying to the Center Of Mass (COM). 

Assuming that the COM position $\mathbf{c}$ is a linear function of the root position $\mathbf{R}$, given user-defined \textbf{symmetric} bounds on the COM dynamics along three orthogonal axis: 
\begin{equation} \label{eq:dimt_bounds}
\begin{aligned}
	-\dcom^{max}_{\{\vect{x},\vect{y},\vect{z}\}}	& \leq & \dcom_{\{\vect{x},\vect{y},\vect{z}\}}	&	\leq & \dcom^{max}_{\{\vect{x},\vect{y},\vect{z}\}} \\
	-\ddcom^{max}_{\{\vect{x},\vect{y},\vect{z}\}}	& \leq & \ddcom_{\{\vect{x},\vect{y},\vect{z}\}} &	\leq & \ddcom^{max}_{\{\vect{x},\vect{y},\vect{z}\}} \,. \\
\end{aligned} 
\end{equation}
Given also an initial state $\vect{S}_0 =  <\vect{R}_0,  \com_{0},\dcom_{0}>$ and a target state $\vect{S}_1 =  <\vect{R}_1, \com_{1},\dcom_{1}>$, the DIMT
outputs a minimum time Bezier curve $\vect{c}(t)$ that connects exactly $<\vect{c}_0,\dcom_{0}>$ and $<\vect{c}_1,\dcom_{1}>$ and satisfies \eqref{eq:dimt_bounds} without considering collision avoidance or the reachability condition. From  $\vect{c}(t)$ we can directly retrieve the translation component of $\vect{R}(t)$, while the orientation is given by interpolating between $\vect{R}_0$ and $\vect{R}_1$. 

However, the DIMT method is not directly applicable as for legged robots the center of mass acceleration bounds are neither constant nor symmetric, but state-dependent. The bounds correspond to the non-slipping condition, and are thus determined by the COM position, as well as the contact points and normals.

To address this issue, we propose a two-step method:
\begin{itemize}
\item We use the DIMT method with acceleration constraints computed for the initial state $\state_0$. The probable contact positions and normals are estimated using the reachability condition. Given the root position $\vect{R}_0$, we compute the set of obstacles intersected by the reachable workspace and arbitrarily choose one position on one of the available surfaces as the plausible contact. By doing so, we increase the odds that the trajectory $\vect{c}(t)$ be dynamically feasible in the neighborhood of $\state_0$, but not along the complete trajectory.
\item Then, in the trajectory validation phase, we verify the dynamic equilibrium of the robot by estimating the probable contact points along the trajectory and verifying the Newton-Euler equation for the COM given this estimate. We also verify the reachability condition (which also approximates collision avoidance constraints). The returned trajectory  $\vect{c}^\prime(t)$ is the part of  $\vect{c}(t)$ that satisfies all these constraints.
\end{itemize}

We refer the reader to the original paper \cite{fernbach2017kinodynamic} for details on how the acceleration bounds can be extracted from a state and estimated positions using the definition of the centroidal wrench cone and empirical evidence of the efficiency of this approach in spite of the heuristics introduced.

\section*{Acknowledgment}

D. Song and Y.-J. Kim are supported in part by the ITRC/IITP program (IITP-2021-0-01460) and the NRF (2017R1A2B3012701) in South Korea. The other authors are supported by the H2020 project Memmo (ICT-780l684).

\ifCLASSOPTIONcaptionsoff
  \newpage
\fi



%



\bibliographystyle{IEEEtran}{}
\bibliography{references}

%








\end{document}